\documentclass{article}

\PassOptionsToPackage{numbers}{natbib}

  \usepackage[preprint]{neurips_2019}

\usepackage[utf8]{inputenc} 
\usepackage[T1]{fontenc}    
\usepackage{hyperref}       
\usepackage{url}            
\usepackage{booktabs}       
\usepackage{amsfonts}       
\usepackage{nicefrac}       
\usepackage{microtype}      
\usepackage{graphicx}
\usepackage{amsmath}
\usepackage{extarrows,xfrac,bm,nameref,dsfont,color,cancel}
\usepackage{wrapfig}

\title{CASS: Cross Adversarial Source Separation via Autoencoder}

\author{%
  Yong~Zheng~Ong \\
  Department of Mathematics \\
  National University of Singapore \\
  10 Lower Kent Ridge Road \\
  Singapore 119076 \\
  \texttt{e0011814@u.nus.edu} \\
  \And
  Charles~Chui\\
  Department of Mathematics\\
   Hong Kong Baptist University\\
    Kowloon, Hong Kong \\
  \texttt{charleskchui@hkbu.edu.hk} \\
  \And
  Haizhao~Yang \\
  Department of Mathematics and Institute of Data Science \\
  National University of Singapore \\
  10 Lower Kent Ridge Road \\
  Singapore 119076 \\
  \texttt{haizhao@nus.edu.sg} \\
}

\begin{document}

\maketitle

\begin{abstract}
  This paper introduces a cross adversarial source separation (CASS) framework via autoencoder, a new model that aims at separating an input signal consisting of a mixture of multiple components into individual components defined via adversarial learning and autoencoder fitting. CASS unifies popular generative networks like auto-encoders (AEs) and generative adversarial networks (GANs) in a single framework. The basic building block that filters the input signal and reconstructs the $i$-th target component is a pair of deep neural networks $\mathcal{EN}_i$ and $\mathcal{DE}_i$ as an encoder for dimension reduction and a decoder for component reconstruction, respectively. The decoder $\mathcal{DE}_i$ as a generator is enhanced by a discriminator network $\mathcal{D}_i$ that favors signal structures of the $i$-th component in the $i$-th given dataset as guidance through adversarial learning. In contrast with existing practices in AEs which trains each Auto-Encoder independently, or in GANs that share the same generator, we introduce cross adversarial training that emphasizes adversarial relation between any arbitrary network pairs $(\mathcal{DE}_i,\mathcal{D}_j)$, achieving state-of-the-art performance especially when target components share similar data structures.
\end{abstract}

\section{Introduction}
\label{sec:introduction}

Source separation refers to the processing of an input signal which is made up of a mixture of multiple component signals into their corresponding components. This problem has numerous applications in a wide range of fields, for example medical electrocardiography (ECG) separation \cite{Ref15}, musical audio separation \cite{Ref14,Ref1,Ref2}, photoplethysmography (PPG) signals \cite{Ref16} and magnetoencephalography (MEG) signals \cite{Ref17}.

Traditional methods like independent component analysis \cite{Ref18}, non-negative matrix factorization \cite{Ref19} have been proposed to solve this problem. Recently, neural networks have seen increasing popularity as a solution to tackling this problem. The use of auto-encoders (AEs) have been proposed as an approach to supervised source separation, by for example \citet{Ref1} and \citet{Ref2}. In these papers, each component of the input signal is captured by an auto-encoder, and training is done on each auto-encoder independently. Generative adversarial networks (GANs) have also been proposed with success. Take for example \citet{Ref4} and \citet{Ref20} who proposed the use of a generator to model the source separation process. In these papers, a single generator is used to model the separation process and generate multiple components, and the outputs trained adversarially using discriminators.

This paper proposes a new model, cross adversarial source separation (CASS) for tackling source separation. CASS aims to unify existing AE and GAN architectures into a single framework. The model builds upon existing AE frameworks like \cite{Ref1} through the introduction of GAN training objectives in each component. Motivated by the idea that in real life applications, source separation tasks involves component signals often of some dependence between each other, CASS introduces cross adversarial training with the help of the additional discriminators to emphasize adversarial relations between each component, as a novel method of information sharing across neural networks, in comparison to existing AE models that trains components independently. 

Cross adversarial training is done on each of the components by, in addition to training the discriminator of each component using standard GAN training methods, also training each discriminator to reject synthesized samples from the other components. Through this method, information about other components can then be relayed back to the AE of each component through backpropagation. The advantage of cross adversarial training is more pronounced on structurally similar components, which we will provide empirical evidence in Section \ref{sec:experiments}. 

Cross adversarial training method also allows discriminators to distinguish structurally similar components, and allows for larger customization of the framework, by controlling the direction of information flow between the components. For example, we provide empirical results to show how it is sometimes desirable to pass on information belonging to a highly represented component to an underrepresented component, but the converse may not be optimal.

CASS also provides a new approach to using GAN in source separation tasks. In contrast to existing GAN models \cite{Ref4,Ref20} which models the separation process with a single generator, CASS proposes the use of multiple generators each tagged to a component and an AE. Instead of using the original mixture spectrum as input to the generator, the Encoder provides a smaller dimension feature representation of the mixture signal, allowing us to work with larger input mixture signals without concern about network size.

This paper will be organized in the following way: Section \ref{sec:prelims} will briefly discuss the building blocks for CASS, AEs and GANs, followed by Section \ref{sec:proposed} which will introduce the model architecture of CASS. Section \ref{sec:experiments} shows some empirical results comparing the proposed model with existing AE and GAN frameworks. Finally, we end with Section \ref{sec:conclusion} which provides a summary of this paper.

\section{Preliminary}
\label{sec:prelims}

\subsection{Auto-Encoders}

An AE framework can be explicitly defined by 2 functions. The first function, $\mathcal{EN}$, is a feature extracting function computing a feature vector $h=\mathcal{EN}(X;\theta_e)$. This function called the encoder encodes input $X$ to some feature representation $h$ of $X$. The other function, $\mathcal{DE}$, is a generating function, and maps a feature vector $h$ back to a reconstructed image $\Bar{X}=\mathcal{DE}(h;\theta_d)$. This function, called the decoder, decodes the feature vector $h$ and attempts to reconstruct $X$, based on the encoded feature vector. $\theta_e$ and $\theta_d$ refers to the weights and biases in the encoder and decoder respectively. Both functions explicitly define an auto-encoder neural network, which attempts to solve
\begin{equation}
    \label{eqn:ae_function}
    \Bar{X}\approx\mathcal{DE}\circ \mathcal{EN}(X;\theta_e,\theta_d).
\end{equation}
Source separation using AEs is done by modelling each source with a single AE network \cite{Ref1}. Given input mixed signal $X$ constituting of $K$ components $X_i$, $i\in\{1,\dots,K\}$, for example, if $X$ is some orchestral music signal, $X_i$ could represent the signals generated by the $i$-th instrument. Each AE, which we denote as $\mathcal{AE}_i$ is then trained to take in an input mixed signal $X$, and output a component signal $\Bar{X_i}$, using the corresponding loss function $L_i(\Bar{X_i},X_i)$ which computes loss between the reconstructed $i$-th component source against the $i$-th component in the training sample.

The learning process for the $i$-th AE can then be described as minimizing a loss function
\begin{equation}
    \label{eqn:ae_loss}
    L(X_i,\mathcal{AE}_i(X)),
\end{equation}
where $L$ is a loss function penalizing differences between the reconstructed $i$-th component and the ground truth $X_i$ \cite{Ref13}, and $\mathcal{AE}_i=\mathcal{DE}_i\circ \mathcal{EN}_i$ represents the $i$-th AE taking as input the mixture spectrum and output the $i$-th component spectrum. For example, in a standard AE model, $L$ could be the Mean-Squared Error.

\subsection{Generative Adversarial Networks}

Generative Adversarial Networks (GANs) are another type of generative model introduced in 2014 by \citet{Ref11}. In contrast to AEs, which attempts to learn an identity mapping of the input space, GAN approaches model generation through the introduction of an adversarial network and attempts to produce output images from some noise distribution in order to fool this adversarial network.

A GAN is made up of two separate network functions. The generator, $G$, takes a random noise vector $z$ from noise distribution $p_z$ and outputs a synthetic sample $G(z;\theta_G)$, where $\theta_G$ refers to the corresponding weight matrices which parameterize the neural network $G$. We denote the output as samples drawn from a distribution $G(z;\theta_G)\sim p_G$. On the other hand, we have ground truth training samples $X$ which are drawn from some data distribution $p_{data}$. The other network, the discriminator, $D$, takes an input $X$ or $G(z)$, and outputs a value $D(X;\theta_D)$ or $D(G(z);\theta_D)\in [0,1]$ denoting the probability of the input being a true sample \cite{Ref12}.

Within this setting, both networks can be trained using the below min-max function
\begin{equation}
    \label{eqn:gan_loss}
    \min\limits_G\max\limits_D\ \mathop{\mathbb{E}}_{X\sim p_{data}}[\log D(X)]+\mathop{\mathbb{E}}_{z\sim p_{z}}[1-\log D(G(z))],
\end{equation}
where $X$ is sampled from the actual data distribution $p_{data}$ and $z$ from some random noise distribution $p_z$. With this objective function, the generator is trained to generate samples which fools the discriminator into thinking that the generated sample is real, and the discriminator is learned to distinguish between real samples from the data distribution versus synthetic fake data from the generator. This way, the generator learns the data distribution $p_{data}$.

\section{Proposed Model}
\label{sec:proposed}

\subsection{Model Outline}

\begin{figure}[h!]
    \centering
    \begin{minipage}{0.3\textwidth}
        \centering
        \includegraphics[width=1\textwidth]{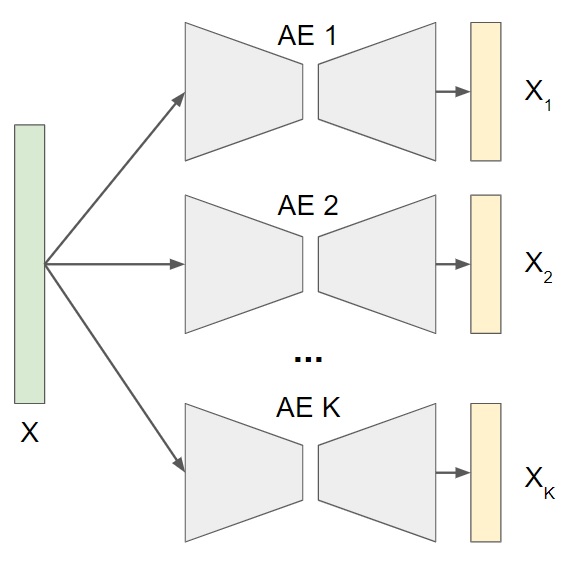}
        \caption{Baseline AE model design}
        \label{fig:baseline}
    \end{minipage}\hfill
    \begin{minipage}{0.33\textwidth}
        \centering
        \includegraphics[width=1\textwidth]{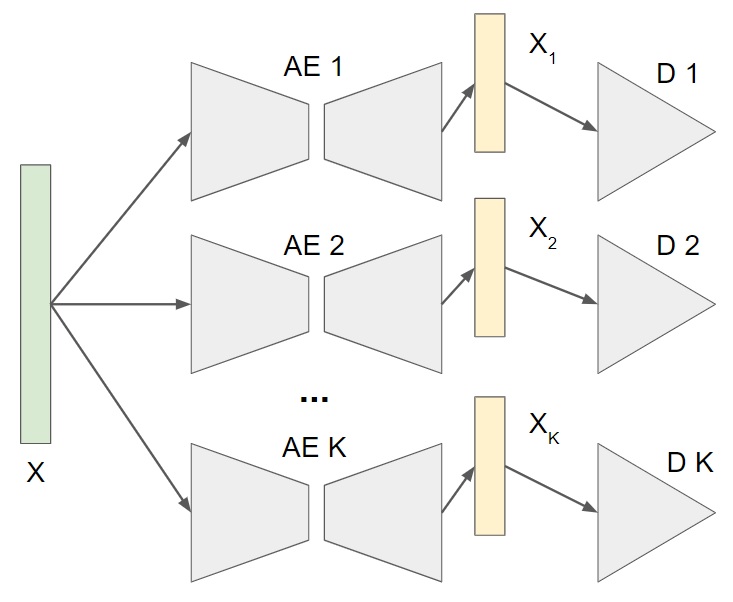}
        \caption{Regular CASS model design}
        \label{fig:cass}
    \end{minipage}\hfill
    \begin{minipage}{0.33\textwidth}
        \centering
        \includegraphics[width=1\textwidth]{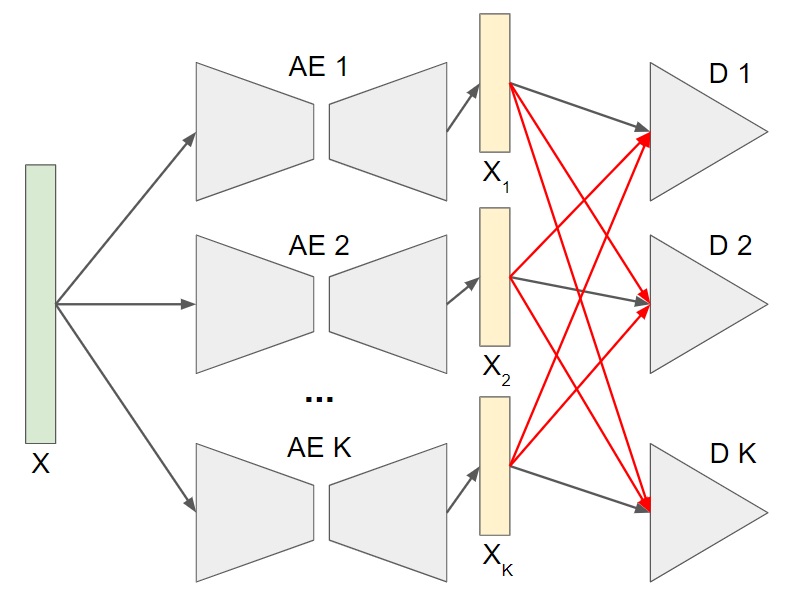}
        \caption{CASS with Cross Adversarial Training}
        \label{fig:cass_c}
    \end{minipage}
\end{figure}

We begin by introducing some notations. Suppose our input signal is denoted by $X$ which is a mixture of $K$ component signals $X_k$. Then, the $i$-th component is modeled using a pair of deep neural networks $\mathcal{EN}_i$ and $\mathcal{DE}_i$ as an encoder for dimension reduction and a decoder for component reconstruction respectively. This network is enhanced using a discriminator network $\mathcal{D}_i$ which favors signal structures of the $i$-th component. The auto-encoder pair $\mathcal{EN}_i$ and $\mathcal{DE}_i$ takes as input a mixture spectrum of the signal, $X$, and outputs the $i$-th component spectrum reconstruction $\bar{X_i}$, similar to Equation \eqref{eqn:ae_function}. The discriminator $\mathcal{D}_i$ then distinguishes between the ground truth $X_i$ and $\bar{X_i}$. Figure \ref{fig:cass} shows the overall design of the regular CASS model, compared to the AE framework used by \citet{Ref1} in Figure \ref{fig:baseline}.

This framework of using GAN in source separation differs from existing methods used in training GAN \cite{Ref3,Ref4,Ref5}. In these applications, a single generator is used to model the source separation task with input as the mixture signal $X$, and outputs all the components either concatenated or separated using a $1 \times 1$ Convolutional Kernel.

\subsection{Training Objective}

Let us denote each network pair $(\mathcal{EN}_i, \mathcal{DE}_i)$ representing the $i$-th encoder and decoder respectively as $\mathcal{AE}_i$, such that the reconstruction of the $i$-th component can be viewed as $\bar{X_i}=\mathcal{AE}_i(X)=\mathcal{DE}_i\circ \mathcal{EN}_i(X)$ as in Equation \eqref{eqn:ae_function}. Training of each $\mathcal{AE}_i$ is done using an AE loss function described by Equation \eqref{eqn:ae_loss}. GAN is used as a way to supplement AE training, following the below loss function in Equation \eqref{eqn:cass_loss}. At training of the $i$-th component, the following loss function is used
\begin{equation}
    \label{eqn:cass_loss}
    \min\limits_{\mathcal{AE}_i}\max\limits_{\mathcal{D}_i}\ \alpha L(X_i, \mathcal{AE}_i(X)) + \beta\{\mathop{\mathbb{E}}_{X_i\sim p_{data,i}}[\log \mathcal{D}_i(X_i)]+\mathop{\mathbb{E}}_{X\sim p_{data,mix}}[1-\log \mathcal{D}_i(\mathcal{AE}_i(X))]\},
\end{equation}
where $p_{data,i}$ refers to the ground truth $i$-th component data distribution, $p_{data,mix}$ refers to the mixture data distribution, $\alpha$ and $\beta$ are tunable parameters which determines the weightage of each component that sums up to 1. In a nutshell, the training objective can be seen as training a weighted sum of AE loss and GAN loss. Training is done by training the AE and discriminator alternately.

\subsection{Cross Adversarial Training}

The above training objective \eqref{eqn:cass_loss} trains each component independently, just like existing AE networks. Training of each component can be summarized as separating of $X_i$ as the actual signal, while the mixture of the other components is interpreted as "noise" in that particular network. However, that is usually not the case in real applications. For example, in music signals, beats of different musical instruments are arranged together, and understanding the structure of another musical instrument in the music could help in providing more information towards the shape of another.

The introduction of the GAN training, specifically the discriminator, provides a method to introduce information sharing between different components. This is done in the following way. Instead of training the discriminator to only reject samples generated from the $i$-th component. The discriminator is also further trained to reject the synthesized samples of each of the other components. This adds the following term to Equation \eqref{eqn:cass_loss}
\begin{equation}
    \sum\limits_{j\neq i}\alpha_j\{\mathop{\mathbb{E}}_{X\sim p_{data,mix}}[1-\log \mathcal{D}_i(\mathcal{AE}_j(X))]\},
    \label{eqn:cass_c}
\end{equation}
where $\mathcal{AE}_j$ refers to each of the other components, and $\alpha_j$ the weightage of the Cross Adversarial Training term for the $j$-th component. $\alpha_j$ controls the extent of information sharing we want for the $i$-th component, hence allowing for control of information flow. Figure \ref{fig:cass_c} shows the additional cross adversarial training, where inputs from other components are also used to train the $i$-th discriminator.

\subsection{Motivation}

The motivation for the design of CASS is as follows. The baseline framework for CASS is extended from the multi-network design used by AEs in Source Separation \cite{Ref1}. This allows the Encoder of each AE to provide a feature representation corresponding to that component from the mixture signal. In comparison, modelling the entire source separation process using a single AE like in GAN results in a single Encoder that compresses the input signal to a general feature representation of the mixed signal in order to extract each component, rather than specific features corresponding to each individual component.

In existing GAN models proposed to tackle source separation, a single generator is used, taking the mixture signal as input, and outputting all $K$ components. Due to the GAN framework, dimensions of the hidden layers would be larger than that of the input signal, and thus if the input signal is long, the resulting network is large and requires a large amount of memory to train. In comparison, the multi-network AE design becomes much more suitable for source separation tasks. Encoder networks provide decompression and feature extraction of the original input signal, such that the size of the decoder is much smaller. Furthermore, splitting each component into separate networks allows each encoder to learn specific feature representations of the original mixture most suited for extracting the $i$-th component. This becomes particularly important when working with a large number of components. Separating the training of each component divides the large network based on the number of components, and training of each component can also be done in parallel.

On the other hand, AE frameworks train each component independently. Compared to GANs which use a single generator to generate all the components, existing GAN frameworks are able to capture dependent information between the different components. To handle this concern, CASS proposes the cross adversarial training that emphasizes adversarial relations between each component, as a means of information sharing using the discriminator $\mathcal{D}_i$. Cross adversarial training allows the discriminator to learn features in the other components, which is transferred to the AE through backpropagation. In addition, we will also show, in Section \ref{sec:experiments}, that in some cases, the transfer of information may not be desirable, e.g. from underrepresented component to more highly represented components. In this case, CASS allows for customization and control over the information sharing process, through introducing cross adversarial training only on components which benefit from this action.

\section{Experiments}
\label{sec:experiments}

This section looks at some real-life applications of the proposed model and compares it to other existing network designs. We consider three applications and analyze the performances of CASS in audio signal separation, electrocardiography (ECG) signals and Photoplethysmography (PPG) signals.

In all the experiments, training is conducted with a learning rate of 0.00001 for the $AE$ and 0.000001 for the discriminator. The batch size used is 50. The discriminator learning rate is lower as the weight of the training of the generator portion is lower. We train using parameters for the $AE$ network with loss weight $\alpha=0.9$ and $\beta=0.1$ in Equation \eqref{eqn:cass_loss}. For the discriminator, cross adversarial training was done with weight $\alpha_j=0.01$. 1 Tesla K40t GPU is used for training each model separately. After every epoch, the error, computed using $l_2$ norm, is recorded. At the end of the training, the results are scored by comparing the results of a test set with their corresponding ground truth components, by computing the relative $p$-norm errors. The relative error of observing $\bar{x}$ from the reconstruction of $x$ is computed on a test set using
\begin{equation}
    error(\bar{x},x)=\frac{||\bar{x}-x||}{||x||}.
\end{equation}
For the following experiments, we compare the performance of CASS with (Figure \ref{fig:cass_c}) and without (Figure \ref{fig:cass}) the additional cross adversarial training against a baseline model. We employ a ResNet-9 architecture for each encoder, decoder and discriminator. This model used is a simplified variation of ResNet-50 from \cite{Ref21}. The baseline model used would be the AE framework used in \cite{Ref1}, with framework in Figure \ref{fig:baseline}. As the objective is to compare the performance of the introduction of a GAN training component, the models will be trained using losses used in standard AEs and GANs, mean squared error for the AE loss, and binary cross entropy for GAN training, instead of losses in more complicated models like variational auto-encoders used in the original paper. That is, the following loss function is used to train the AE in the $i$-th AE
\begin{equation}
    \label{eqn:ae_loss}
    \alpha*MSELoss(\mathcal{AE}_i(X),X_i)+\beta*BCELoss(\mathcal{D}_i(\mathcal{AE}_i(X)),1),
\end{equation}
where $MSELoss$ and $BCELoss$ refer to the mean squared error and binary cross entropy Error respectively, and $\alpha$ and $\beta$ are as described above. The $i$-th discriminator is trained using
\begin{equation}
    \label{eqn:d_loss}
    BCELoss(\mathcal{D}_i(\mathcal{AE}_i(X)),0)+BCELoss(\mathcal{D}_i(X_i),1),
\end{equation}
while CASS with cross adversarial training trains the $i$-th discriminator using the following loss function
\begin{equation}
    BCELoss(\mathcal{D}_i(\mathcal{AE}_i(X),0)+BCELoss(\mathcal{D}_i(X_i),1)+\sum\limits_{j\neq i}\alpha_jBCELoss(\mathcal{D}_i(AE_j(X)),0),
\end{equation}
where the additional term emphasizes adversarial relations between any arbitrary component.

\subsection{Dataset}

Dataset for audio signal separation is obtained from the Bach10 dataset \cite{Ref6}. The dataset contains audio recordings of each musical instruments and the ensemble of these musical instruments. The audio recordings of two instruments, the bass and saxophone, is considered for separation and the objective is then to separate both instruments from the ensemble of them.

In ECG signal separation, the objective is to separate fetal ECG and maternal ECG from abnormal ECG signals containing a mixture of both, and external respiratory noise. Due to the lack of ground truth component signal datasets, and the objective of this application is to compare structurally similar components, synthetic data is used for this experiment to highlight these properties in our dataset, and is generated using Python package signalz \cite{Ref7} for the maternal and fetal ECG. During pregnancy, cardiac output of the mother increases by 30-50\%, and as a result, average beats per minute (BPM) of a pregnant mother is about 80 to 90 \cite{Ref8}. On the other hand, the estimated average fetal BPM ranges from 120 to 160 \cite{Ref9}. Furthermore, maternal ECG magnitude measured from the abdominal signal is 2 - 10 times that of the fetal ECG signal \cite{Ref10}. Thus, synthetic samples of uniformly generated BPM along the above ranges and magnitudes were generated and mixed. Noise is introduced using a random sinusoidal wave with varying frequency and amplitude to signify periodic respiratory noise.

In PPG signal separation, the objective is to separate heart beat PPG signals and respiratory PPG signals. The samples were synthetically generated using randomly generated BPM for heart beats and breathing frequency. Reference samples for heart beat signals and respiratory signals were obtained from \cite{Ref22}. The purpose of this experiment is to investigate the effect of Cross Adversarial Training in less related components, so to get more accurate comparisons, data is modelled using PPG signals obtained at rest, with minimal motion artifact noise.

For all our applications, original mixture signals are first preprocessed using the short time Fourier transform (STFT) and the spectrogram used as inputs for training.

\subsection{Audio Signal Separation}

\begin{wrapfigure}{r}{0.5\textwidth}
    \includegraphics[width=0.48\textwidth]{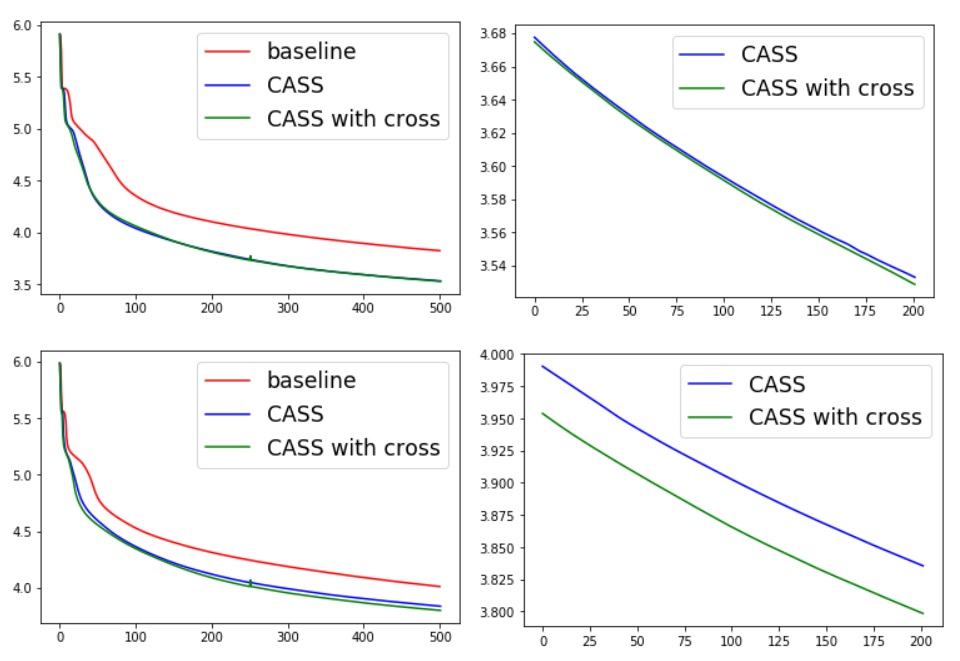}
    \caption{log of testing error for audio separation during training of 500 epochs. Top: error for the bass network. Bottom: error for the sax network. Left: full error for training all epochs. Right: error of last 200 epochs.}
    \label{fig:audio_l2error}
\end{wrapfigure}

Figure \ref{fig:audio_l2error} shows the error observed during training. It can be seen that the addition of the GAN training objectives in CASS does improve the performance of the neural network as compared to the baseline AE. The right figure shows the errors of the last 200 epochs of the training data for CASS with and without the cross adversarial training objective. Clearly, the sharing of information improves the absolute error during training slightly.

Comparing the computed $p$-norm errors after inverse STFT is performed on reproduced signals in Table \ref{tab:audio_error}, we can see that the relative errors with cross adversarial training are slightly better than without information sharing for both $L_1$ and $L_2$ norms. Furthermore, both proposed models perform much better than the baseline AE. We argue that this is because the additional information shared by the other component provided additional information for the AE in each component to learn the relations between the structures of the source signals.

\begin{table}[h!]
    \centering
    \begin{tabular}{llll}
        \hline
        \textbf{Bass Model} & \textbf{$L_1$ Error} & \textbf{$L_2$ Error} & \textbf{$L_\infty$ Error}\\ \hline
        Baseline AE & 0.13942 & 0.14630 & 0.21712 \\
        CASS & 0.10363 & 0.10909 & 0.16692 \\
        CASS with Cross Training & 0.10241 & 0.10825 & 0.16887
    \end{tabular}
    \begin{tabular}{llll}
        \hline
        \textbf{Saxophone Model} & \textbf{$L_1$ Error} & \textbf{$L_2$ Error} & \textbf{$L_\infty$ Error}\\ \hline
        Baseline AE & 0.15845 & 0.16984 & 0.26421 \\
        CASS & 0.13325 & 0.14254 & 0.22329 \\
        CASS with Cross Training & 0.12850 & 0.13717 & 0.21234
    \end{tabular}
    \caption{Different relative $p$-norm errors after training of Bach10 dataset.}
    \label{tab:audio_error}
\end{table}

\subsection{ECG Signal Separation}

This experiment aims to investigate the effect of information sharing on highly similar data, where one component is largely represented and the other underrepresented, and often masked by large noise. General ECG signals follow a standard shape, consisting of 3 components. The $P$ segment, the $QRS$ complex followed by a $T$ wave. Thus, clean maternal and Fetal ECG samples have large amounts of similarities, while their main difference lies in the BPM and the amount of representation of the two signals. Maternal ECG signals are more highly present from abdominal signals, whereas the magnitude of fetal ECG is small, and often masked by a noise like a respiratory signal. The highly similar data, and different magnitudes of components, when mixed together, result in a challenging dataset to train on.

Table \ref{tab:ecg_error} records the relative $L_1$, $L_2$ and $L_\infty$ errors computed on a testing set at the end of the training. On the maternal ECG data, it can be observed that the model performs best on CASS model without any cross adversarial training. Cross adversarial training actually performs worse than the model without, but still, both models perform better compared to the baseline AE. We argue that this is because the maternal ECG signal is the signal which is more represented in the input mixture data. Maternal ECG magnitude measured from the abdominal signal is 2 - 10 times that of the fetal ECG signal \cite{Ref10}. Being the prominent signal, performing a cross training on Fetal ECG signals would, instead of improving the performance, lead to confusion of the discriminator. Overall, however, the introduction of GAN training still enhances the performance of the model compared to that of the baseline AE.

\begin{table}[h!]
    \centering
    \begin{tabular}{llll}
        \hline
        \textbf{Maternal Model} & \textbf{$L_1$ Error} & \textbf{$L_2$ Error} & \textbf{$L_\infty$ Error}\\ \hline
        Baseline AE & 0.45158 & 0.53502 & 0.85077 \\
        CASS & 0.40672 & 0.47942 & 0.77370 \\
        CASS with Cross Training & 0.40994 & 0.48082 & 0.77911
    \end{tabular}
    \begin{tabular}{llll}
        \hline
        \textbf{Fetal Model} & \textbf{$L_1$ Error} & \textbf{$L_2$ Error} & \textbf{$L_\infty$ Error}\\ \hline
        Baseline AE & 0.49818 & 0.57435 & 0.80709 \\
        CASS & 0.37387 & 0.46627 & 0.75402 \\
        CASS with Cross Training & 0.37218 & 0.45848 & 0.74462
    \end{tabular}
    \caption{Different relative $p$-norm errors after training of ECG dataset.}
    \label{tab:ecg_error}
\end{table}

\begin{figure}[h!]
    \centering
    \begin{minipage}{0.45\textwidth}
        \centering
        \includegraphics[width=0.9\textwidth]{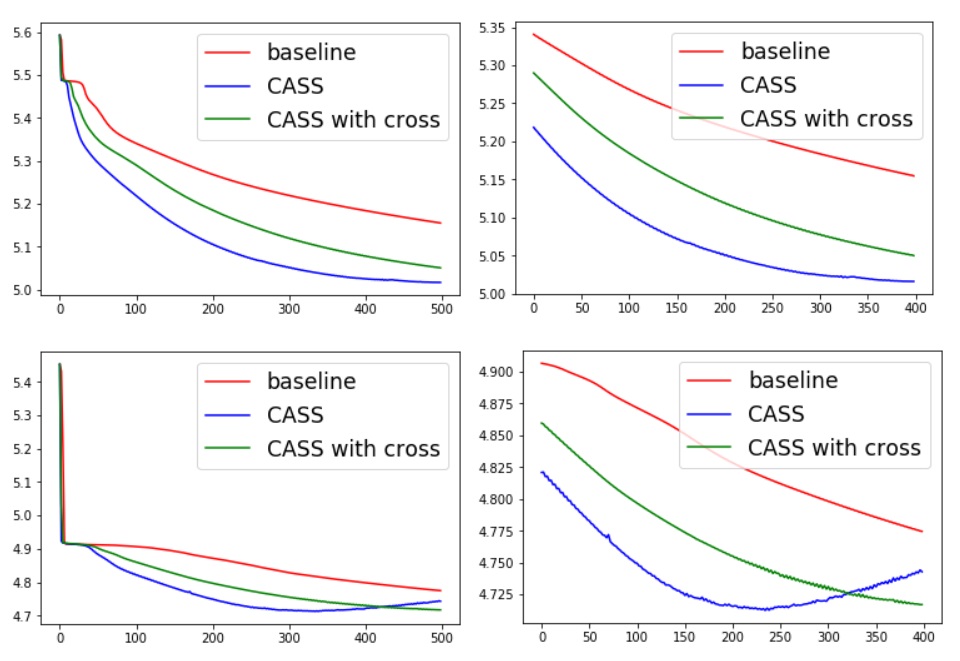}
        \caption{log of testing error for ECG separation during training of 500 epochs. Top: error for the maternal network. Bottom: error for the fetal network. Left: full error for training all epochs. Right: error of last 400 epochs.}
        \label{fig:ecg_l2error}
    \end{minipage}\hfill
    \begin{minipage}{0.45\textwidth}
        \centering
        \includegraphics[width=0.9\textwidth]{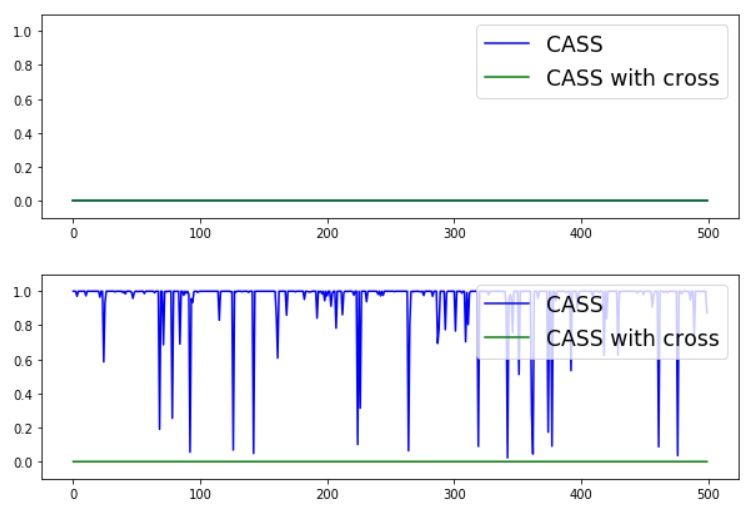}
        \caption{Discriminator output for cross training of data. $x$-axis represents sample number. Top: Output for the maternal network. Bottom: Output for the fetal network.}
        \label{fig:ecg_cross_result}
    \end{minipage}
\end{figure}

While it is true that CASS without cross training performs well on the highly represented maternal ECG data, performance in the underrepresented Fetal ECG signal is poorer. The errors computed for the signals are worse for all of the 3 types of errors compared to cross training. Cross adversarial training performs better, and we suspect that this is because of the information gained from the major maternal ECG signal. Being structurally similar, the fetal ECG, which is under-represented can learn structural details which are similar in the maternal ECG. This allows for fetal ECG, whose magnitude is smaller and hence likely to be confused by noise, to converge. As seen in Figure \ref{fig:ecg_l2error}, although the convergence of the cross adversarial training is slower than without, training progresses stably. In comparison, testing error on training with regular CASS is unstable and starts to increase after about 300 epochs. This is potentially a sign of mode collapse in GAN, hence poor generalization on testing data.

\subsubsection{Effect of Cross Training on Discriminators}

This section compares the impacts of cross training on discriminators. For this experiment, the outputs of the AE for each component is passed into the other component's discriminator and the output values of the discriminator are compared. Figure \ref{fig:ecg_cross_result} shows the output of the discriminator ($y$-axis) against the sample number ($x$-axis). The top figure shows the output of Fetal AE into the maternal discriminator. As the over-represented signal, both CASS with or without cross adversarial training manages to identify the signal as fake signals. However, in the bottom figure, the regular CASS without cross adversarial training mispredicts most of the synthesized maternal AE outputs as real data signals. In comparison, cross adversarial training is able to identify synthesized maternal component signals belonging to the other components, even in the under-represented fetal discriminator. 

This provides empirical evidence that the discriminator with cross adversarial training managed to learn the differentiating features and similarities of the 2 components, behaving similarly to a classifier. This experiment highlights the difference in what the discriminator learns with and without cross adversarial training. In simple terms, with cross adversarial training, besides the discriminator learning to identify real and fake samples, the model also learns to identify components. The regular CASS only learns to distinguish real and fake samples but was unable to learn the differences between the two structurally similar components in the underrepresented model.

\subsection{PPG Signal Separation}

This section compares the effect of cross training on less related data. In separation of PPG signals, heart beat PPG and respiratory PPG exhibit different shapes. Table \ref{tab:ppg_error} records relative errors during the end of the training, computed over a testing set. Again, baseline error for both components perform worse in terms of relative errors compared to proposed CASS models. However, errors in CASS with Cross Adversarial Training actually performs slightly worse than without. we argue that the effect of information sharing on components which are less related instead could worsen overall performance. In this case, the sharing of information between the two components in PPG separation leads to each AE being confused by the outputs of the other component. This highlights that the choice of how information is shared between components plays a role in the performance of each component. CASS provides a flexible way for controlling information flow between components, through varying the values of $\alpha_j$ in Equation \ref{eqn:cass_c}, which would not be possible if a single Network is used to model the Source Separation process like in existing GAN frameworks \cite{Ref3,Ref4,Ref5}.
\begin{table}[h!]
    \centering
    \begin{tabular}{llll}
        \hline
        \textbf{Heart Rate Model} & \textbf{$L_1$ Error} & \textbf{$L_2$ Error} & \textbf{$L_\infty$ Error}\\ \hline
        Baseline AE & 0.02409 & 0.026174 & 0.051244 \\
        CASS & 0.019765 & 0.021612 & 0.04225 \\
        CASS with Cross Training & 0.02047 & 0.022261 & 0.043220
    \end{tabular}
    \begin{tabular}{llll}
        \hline
        \textbf{Respiratory Model} & \textbf{$L_1$ Error} & \textbf{$L_2$ Error} & \textbf{$L_\infty$ Error}\\ \hline
        Baseline AE & 0.05874 & 0.066493 & 0.13064 \\
        CASS & 0.05117 & 0.058100 & 0.11460 \\
        CASS with Cross Training & 0.05310 & 0.06036 & 0.119324
    \end{tabular}
    \caption{Different relative $p$-norm errors after training}
    \label{tab:ppg_error}
\end{table}

\section{Conclusion}
\label{sec:conclusion}

This paper introduces CASS as an alternative approach to source separation by deep learning. Numerical results in Section \ref{sec:experiments} showed that CASS outperforms existing AE models using similar network designs. Furthermore, we proposed Cross Adversarial Training as a solution to information sharing across multiple AE networks and showed that the sharing of information is useful to improve performance of independent networks in cases where there are similar features or under-represented data. PPG example shows the importance of controlling information flow, where CASS provides an edge against existing architectures which either use a single network, thus having no control over information flow between components, or trains each component independently. The code will be availalbe in the authors' personable homepages.



\small
\bibliographystyle{abbrvnat}
\bibliography{refs}   




\end{document}